\documentclass[conference]{IEEEtran}

\usepackage{cite}

\ifCLASSINFOpdf
  \usepackage[pdftex]{graphicx}
\else
\fi

\usepackage{amsmath}

\usepackage{fancyhdr}

\begin{document}

\title{Underwater robot guidance, navigation and \\control in fish net pens}

\author{\IEEEauthorblockN{Sveinung Johan Ohrem}
\IEEEauthorblockA{Aquaculture Robotics and Automation\\ SINTEF Ocean \\ Trondheim, Norway\\
Email: sveinung.ohrem@sintef.no}
}
\maketitle

\begin{abstract}
Aquaculture robotics is receiving increased attention and is subject to unique challenges and opportunities for research and development. Guidance, navigation and control are all important aspects for realizing aquaculture robotics solutions that can greatly benefit the industry in the future. Sensor technologies, navigation methods, motion planners and state control all have a role to play, and this paper introduces some technologies and methods that are currently being applied in research and industry before providing some examples of challenges that can be targeted in the future.
\end{abstract}

\IEEEpeerreviewmaketitle

\fancypagestyle{withfooter}{
  \renewcommand{\headrulewidth}{0pt}
  \fancyfoot[C]{\footnotesize Accepted to the IEEE IROS workshop on Autonomous Robotic Systems in Aquaculture: Research Challenges and Industry Needs}
}
\thispagestyle{withfooter}
\pagestyle{withfooter}

\section{Introduction}
\begin{figure}
    \centering
    \includegraphics[width=\linewidth]{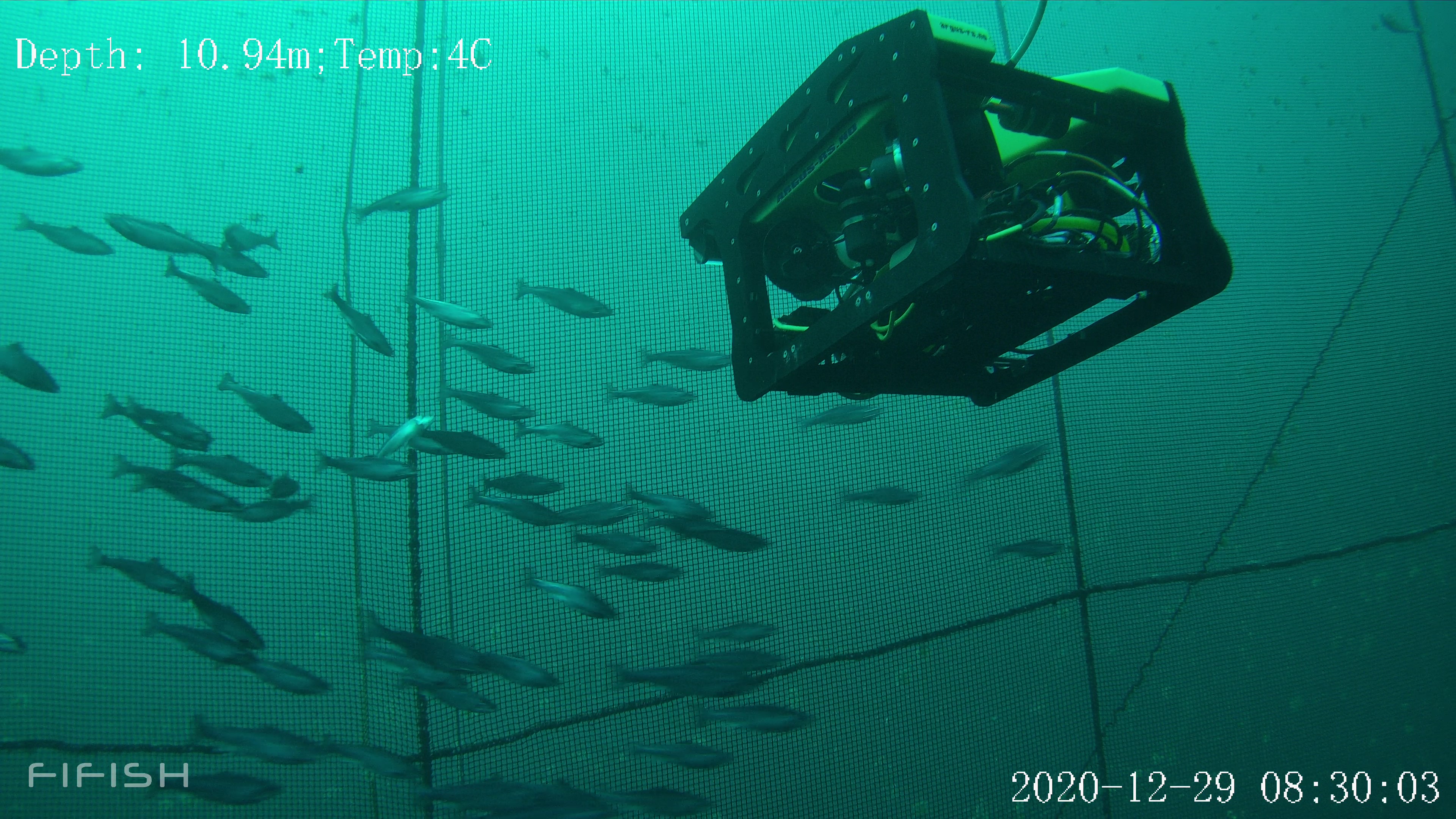}
    \caption{An ROV operating inside an aquaculture net pen. Image: Herman Biørn Amundsen.}
    \label{fig:rov_in_pen}
\end{figure}

The advent of underwater robots in the sea-based aquaculture industry, Fig.~\ref{fig:rov_in_pen}, has motivated the development of various methods for guidance, navigation and control in order to advance the level of autonomy. Operating robots inside a fish net pen is challenging for pilots which experience conflicting tasks during inspection operations: Safely, but quickly maneuver the robot through the net pen while simultaneously inspecting the underwater structures. The pilots can benefit from automatic control functions, but the remotely operated vehicles (ROVs) rarely have other functions than automatic depth and heading control by default. More advanced guidance and control functions can assist the pilot and make their job easier by for instance taking over the control of the vehicle during routine motion executions. However, these advanced methods require methods for navigation inside fish net pens. This is possible if sensors such as Doppler velocity loggers (DVLs) or Ultra-Short Baseline (USBL) systems are employed. The drawback of these sensors is that they are quite costly. Navigation methods utilizing sensors already installed on the ROVs, such as cameras, IMUs and depth sensors would therefore be beneficial.

This paper presents a non-exhaustive overview of various sensor technologies and methods for navigation inside fish net pens and introduces some concepts for guidance and control that have been tested in realistic fish net pen environments. The paper then provides some suggestions for potential future work in aquaculture robotic navigation, guidance and control.

\section{Navigation}
\subsection{Doppler velocity logger (DVL)}
The DVL is commonly used to measure the velocity relative to the sea bottom and the altitude of a vehicle moving either on the sea surface or under water. Hence the DVL is commonly pointed downwards. In an aquaculture setting however~\cite{rundtop2016experimental} presented a solution where the DVL is instead pointed forwards. Thus the DVL provides the vehicles velocity and distance to the net pen structure. This feature is quite useful as the net structure is commonly what is being inspected and it is a useful reference for pilots maneuvering inside a net pen. 

\subsection{Ultra short baseline (USBL)}
\begin{figure}
    \centering
    \includegraphics[width=\linewidth]{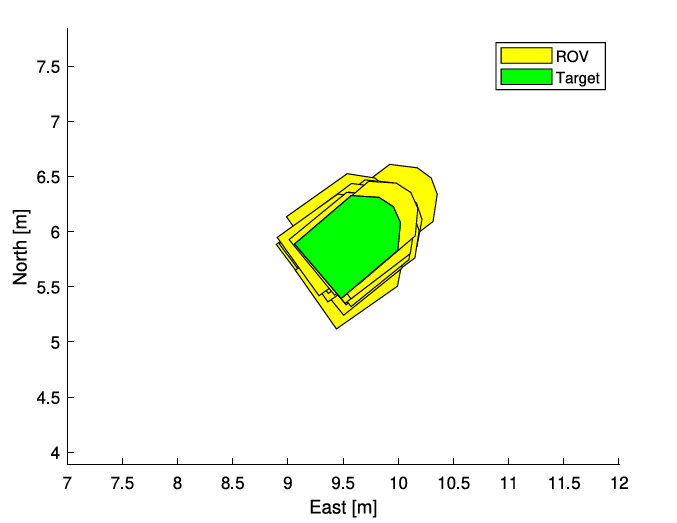}
    \caption{DP footprint of an ROV using a USBL system for positoin measurements. Figure from~\cite{ohrem2022robust} and used under Creative Commons CC-BY license.}
    \label{fig:dp_footprint}
\end{figure}

The USBL technology, sometimes referred to as \emph{underwater GPS} consists of one or more transceivers located at fixed positions and a transponder mounted on the vehicle. This system is used to measure the transponders, hence the underwater vehicles, position relative to the transceiver. The transceiver's position can be fixed in the world frame through global navigation system coordinates, giving the underwater vehicle position in the world frame. In an aquaculture setting the global position of the underwater vehicle is often not required as the vehicle is inside a net pen and will not move over large distances. The vehicle's position relative to the transceiver, however, can be of interest if for instance dynamic positioning (DP) systems are required. A USBL was used for DP inside an aquaculture net pen in~\cite{ohrem2022robust}, Fig.~\ref{fig:dp_footprint}. DP in net pens is useful if the underwater vehicle is monitoring a crowding operation (gathering the fish for e.g. transportation), other net structure operations, or when performing maintenance with a manipulator.

\subsection{Camera-based navigation}
\begin{figure}
    \centering
    \includegraphics[width=\linewidth]{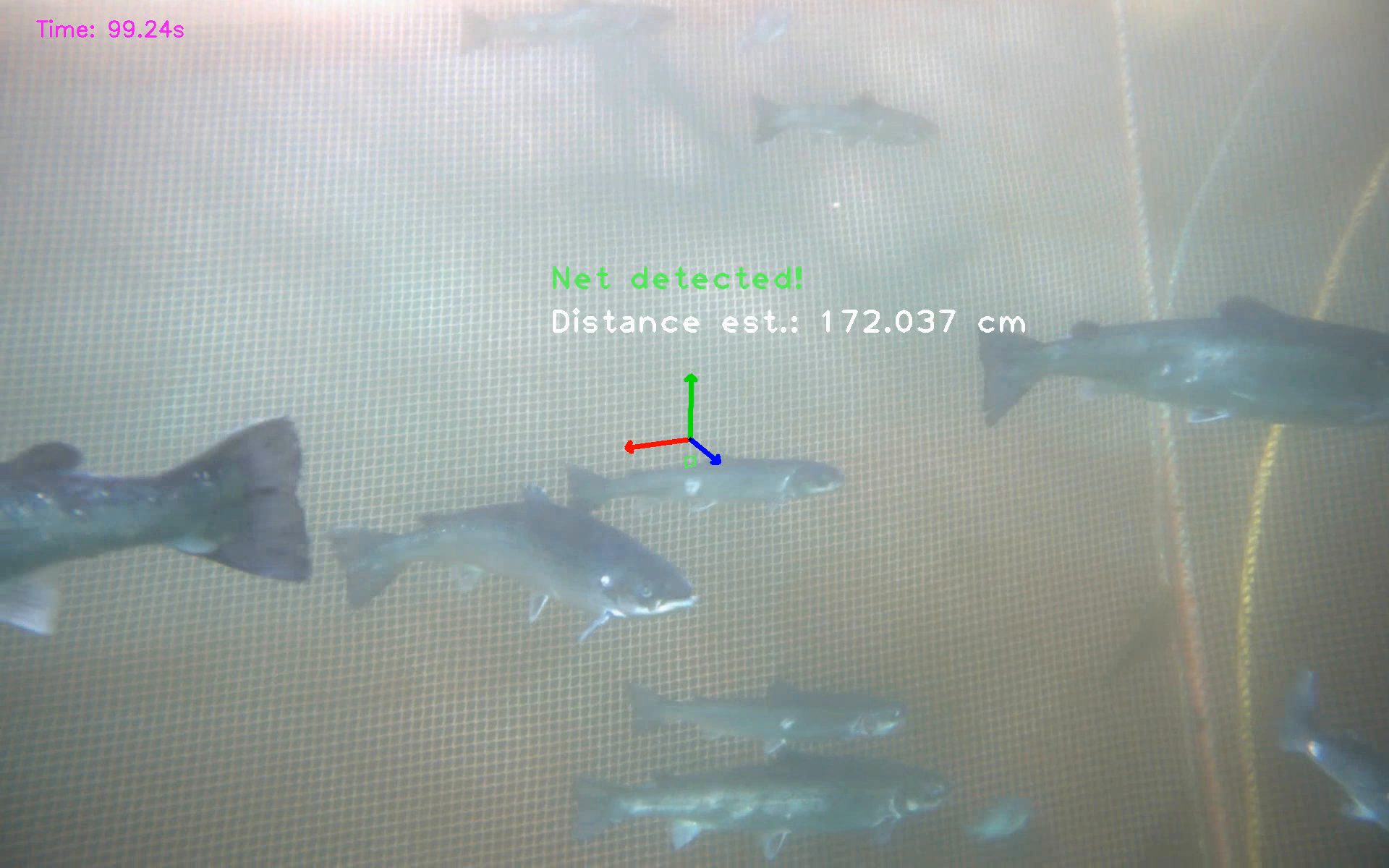}
    \caption{Fast Fourier Transform used to estimate distance and pose of an ROV relative to a net structure.}
    \label{fig:fft}
\end{figure}

The DVL and USBL are costly instruments and underwater vehicles are rarely equipped with these sensors by default. Most vehicles do have a camera, hence using the camera as a sensor for navigation is useful. The camera itself can only capture images, so image processing software is necessary to extract useful information such as vehicle distance to the net, orientation relative to the net and velocity relative to the net. Various computer vision algorithms have been proposed for pose estimation of underwater vehicles in net pens.~\cite{schellewald2021vision} uses 2D Fast Fourier Transform (FFT) to estimate the vehicle distance to the net.~\cite{skaldebo2024approaches} proposes a Fourier Transform method and a stereo camera method to estimate pose and distance, Fig.~\ref{fig:fft}.~\cite{akram2022visual} use vision triangulation methods to identify features on the net structure and use these for navigation.~\cite{bjerkeng2021rov} proposes a laser-camera system for triangulation and net-relative navigation.

\subsection{Local versus global navigation}
When operating in an aquaculture setting the underwater vehicles are often employed inside the net pen structures. This is because the inside of the pens have less ropes and mooring lines in which the vehicle tether might get entangled. There are still very few, if any, vehicles operating without a tether in aquaculture pens. 

Since the vehicle is operating inside the pens which may move and change shape with varying environmental conditions, and is in the grander scales of things a small, known geometric region, it makes sense to also localize and navigate on a smaller, local scale. The positioning systems depending on Global Navigation Satellite Systems (GNSS), or positioning using USBL which gives a position relative to a fixed point, i.e., the transceiver location, not necessarily the structure may not capture the changes in the structure shape.

DVLs and camera-based methods offer localization and navigation relative to detected structures. Hence these navigation methods enables the use of autonomous net-relative motion planning and motion control systems. Further, one may reduce the degrees of freedom in the equations of motion of an underwater vehicle if the operating space is transformed from the global frame (commonly referred to as the North-East-Down frame~\cite{SNAME1950}) to a cylindrical frame, see e.g.~\cite{ohrem2021control,skaldebo2023framework}, where the vehicles distance from the net, the angular position and the depth are instead used to define the position.

\begin{figure}
    \centering
    \includegraphics[width=0.9\linewidth]{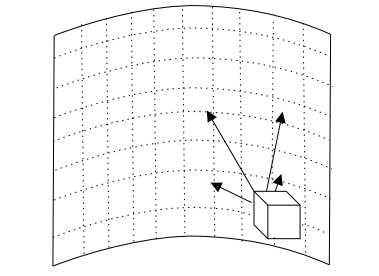}
    \caption{An ROV with a DVL pointed towards the net structure providing distance, heading and velocity measurements. Figure from~\cite{amundsen2021autonomous} and used under Creative Commons CC-BY license}
    \label{fig:rov_dvl}
\end{figure}

\section{Guidance}
Once the vehicle can be localized inside the net pen and it is able to navigate, guidance methods can be employed. One such method is called \emph{net following} and has been demonstrated in various papers~\cite{amundsen2021autonomous,haugalokken2023adaptive,ohrem2023adaptive, karlsen2021autonomous}. The net following algorithm provides the vehicle with a distance, heading and velocity setpoint, all relative to the detected net structure. The algorithm thus requires measurements of these states and the mentioned works all utilize a DVL for this purpose.

\begin{figure}
    \centering
    \includegraphics[width=\linewidth]{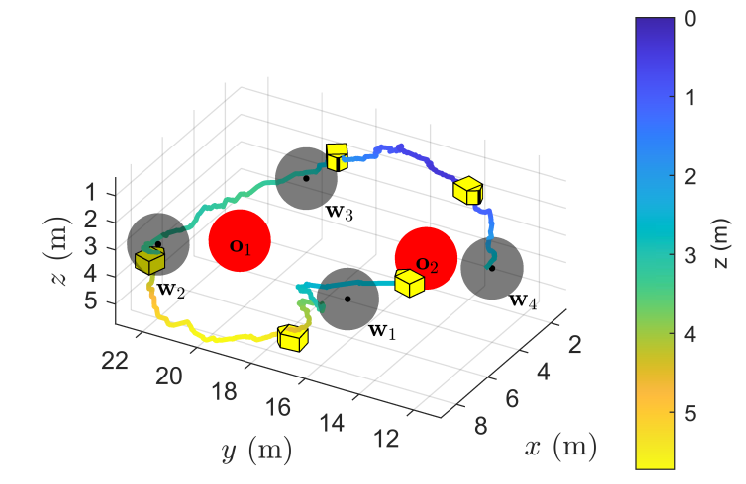}
    \caption{Trajectory of an ROV performing waypoint tracking and avoiding obstacles using the elastic band method. Figure from~\cite{amundsen2024three} used under the Creative Commons CC-BY license.}
    \label{fig:elastic_band}
\end{figure}

If the vehicle is to perform maneuvers that are not relative to the net structure it will likely require a real-time motion planner for guidance, obstacle avoidance an re-planning, alongside navigation tools offering non-structure based navigation (e.g. USBL). Such motion planners have been developed and tested in an aquaculture setting in~\cite{amundsen2024three} where the elastic band approach is used to demonstrate vehicle obstacle avoidance inside a live fish net pen, Fig.~\ref{fig:elastic_band}. Other methods allowing real-time motion planning for underwater vehicles include~\cite{xanthidis2023resiplan, amundsen2024rump}.
\section{Control}
Underwater vehicles have some automatic functions by default. Normally these are limited to control of the states which can be considered trivial to measure, i.e., depth and heading, leaving control of the other states up to the pilot. The roll and pitch angles, however, are in most cases passively stabilized through the vehicle design and thus rarely controlled. Depending on the control objective the North and East positions, the surge, sway and heave speeds and the yaw rate may all require automatic control (the naming convention of the Society of Naval Architects and Marine Engineers (SNAME) is used~\cite{SNAME1950}).

Automatic control of the surge, sway and heave speeds appear to be quite beneficial when operating underwater robots in an aquaculture net pen. Position control is also very helpful if the vehicle is required to e.g. hold a position for long-term inspections. Low-level control methods for surge, sway and heave speed control have been investigated in~\cite{amundsen2021autonomous, karlsen2021autonomous, ohrem2023adaptive, haugalokken2023adaptive, ohrem2024adaptive} and the consensus is that classical PID controllers are outperformed by non-linear and/or adaptive controllers.

\section{Future work and challenges}
\subsection{Navigation}
Cost is the essential keyword here. Both DVL and USBL systems are costly relative to the price of low-cost ROV systems. Cameras however are cheap and already installed on the vehicles. Thus demonstrating that camera-based navigation methods can be applied for navigation inside fish net pens in real-time, under realistic conditions with varying light and visibility is a challenge that when solved is likely to have a potentially huge impact on aquaculture robotics.

\subsection{Guidance}
Some methods have been demonstrated in field trials under realistic conditions, but larger-scale trials, lasting e.g. several days are yet to be performed. The challenge here lies in the development of resilient guidance methods that will enable autonomous operation for underwater vehicles in fish farms.

\subsection{Control}
Control of the vehicle's position and velocity states in an aquaculture setting has been demonstrated in several research papers (e.g. papers mentioned above). These papers, however, lack long-term trials. Such trials could reveal shortcomings in the proposed control methods. The underwater robots employed in aquaculture today are mostly used for inspection. There is a huge potential in also performing maintenance and repair. Thus, developing control methods for vehicle-manipulator systems will have a potentially huge impact on the use of robotics in aquaculture. One immediate challenge is related to performing dynamic positioning relative to objects or features of interest, such as holes in the net, while performing manipulation tasks.

\section{Conclusion}
Navigation, guidance and control of aquaculture robots has seen a significant development over the past years. This paper has provided a brief introduction and overview of some concepts and technologies used for navigation, some examples of guidance methods, and some examples of control before presenting some suggestions for future work and challenges.





\bibliographystyle{IEEEtran}
%


\bibliography{ifacconf}

\end{document}